\documentclass[conference]{IEEEtran}
\IEEEoverridecommandlockouts
\usepackage{cite}
\usepackage{amsmath,amssymb,amsfonts}
\usepackage{algorithmic}
\usepackage{graphicx}
\usepackage{textcomp}
\usepackage{xcolor}
\usepackage{tikz}
\usepackage{placeins}
\usetikzlibrary{arrows,shapes,positioning,calc}
\def\BibTeX{{\rm B\kern-.05em{\sc i\kern-.025em b}\kern-.08em
    T\kern-.1667em\lower.7ex\hbox{E}\kern-.125emX}}
\begin{document}

\title{Dementia Detection using Multi-modal Methods on Audio Data\\
}

\author{\IEEEauthorblockN{\textbf{Saugat Kannojia}}
\IEEEauthorblockA{\textit{IIT Kanpur} \\
saugatkannojia@gmail.com}
\and
\IEEEauthorblockN{\textbf{Anirudh Praveen}}
\IEEEauthorblockA{\textit{IIT Kanpur} \\
anirudhp24@iitk.ac.in}
\and
\IEEEauthorblockN{\textbf{Danish Vasdev}}
\IEEEauthorblockA{\textit{IIT Kanpur} \\
danishvasdev@gmail.com}
\and
\IEEEauthorblockN{Saket Nandedkar}
\IEEEauthorblockA{\textit{IIT Kanpur} \\
Saketsunil23@iitk.ac.in}
\and
\IEEEauthorblockN{Divyansh Mittal}
\IEEEauthorblockA{\textit{IIT Kanpur} \\
divmittal03@gmail.com}
\and
\IEEEauthorblockN{Sarthak Kalankar}
\IEEEauthorblockA{\textit{IIT Kanpur} \\
sarthakk281103@gmail.com}
\and
\IEEEauthorblockN{Shaurya Johari}
\IEEEauthorblockA{\textit{IIT Kanpur} \\
shauryaj23@iitk.ac.in}
\and
\IEEEauthorblockN{Dr. Vipul Arora}
\IEEEauthorblockA{\textit{Dept. of EE, IIT Kanpur} \\
vipular@iitk.ac.in}
}

\maketitle

\begin{abstract}
Dementia is a neurodegenerative disease that causes gradual cognitive impairment, which is very common in the world and undergoes a lot of research every year to prevent and cure it. It severely impacts the patient's ability to remember events and communicate clearly, where most variations of it have no known cure, but early detection can help alleviate symptoms before they become worse. One of the main symptoms of dementia is difficulty in expressing ideas through speech. This paper attempts to talk about a model developed to predict the onset of the disease using audio recordings from patients. An ASR-based model was developed that generates transcripts from the audio files using Whisper model and then applies RoBERTa regression model to generate an MMSE score for the patient. This score can be used to predict the extent to which the cognitive ability of a patient has been affected. We use the PROCESS\_V1 dataset for this task, which is introduced through the PROCESS Grand Challenge 2025. The model achieved an RMSE score of 2.6911 which is around 10 percent lower than the described baseline.
\end{abstract}

\section{Introduction}
A great amount of research has been put into the field of Audio, Speech and Signal Processing to solve real-life problems, majorly focusing on prevention and early detection of diseases in the domain of healthcare. Machine Learning techniques aim to solve such problems with the state-of-the-art techniques to help prevent the onset of diseases like dementia which leads to a gradual cognitive decline which could be delayed or prevented by the timely intervention by the required medical facilities. The organising committee of \textit{Prediction and Recognition Of Cognitive declinE through Spontaneous Speech} (PROCESS) Signal Processing Grand Challenge, aims at detecting various phases of dementia via speech output in the form of audio data by conducting regression and classification tasks.

\section{Dataset}
The dataset provided by PROCESS Grand Challenge committee, labeled PROCESS\_V1, consists of audio recordings for 157 patients that have been diagnosed into one of the three categories. \\
The Healthy Control (HC) group consists of participants who have not been diagnosed with any cognitive impairment, along with those who have shown signs of having memory problems but not due to neurodegenerative diseases. The second category has been decided to be the patients that have gone through Mild Cognitive Impairment (MCI) which is suggestive of early stages of dementia. The third category refers to those participants who have been diagnosed with dementia.\\
Three kinds of tasks were provided to the participants where each of them had to answer a particular question.
\begin{itemize}
    \item Semantic Fluency Task: The question being asked here was “Please name as many animals as you can in a minute.”
    \item Phonemic Fluency Task:  The question being asked here was “Please say as many words beginning with the letter ‘P’ as you can. Any word beginning with ‘P’ except for names of people such as Peter, or countries such as Portugal.”
    \item Cookie Theft Picture Description: The participants were provided with an image as shown in Figure 1, which had to be described when prompted.
    \begin{figure}
        \centering
        \includegraphics[width=0.8\linewidth]{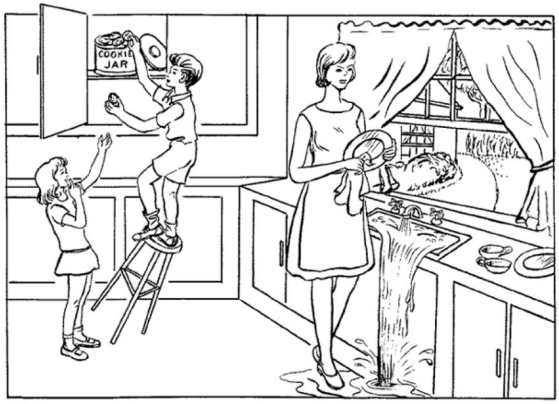}
        \caption{Cookie Theft Picture}
        \label{fig:1}
    \end{figure}
\end{itemize}

\section{Methodology}


We propose a deep learning approach for predicting Mini-Mental State Examination (MMSE) scores from speech recordings. Our approach leverages transfer learning by combining two state-of-the-art models: Whisper \cite{radford2022whisper} for Automatic Speech Recognition (ASR) and RoBERTa \cite{liu2019roberta} for contextual text representation. The pipeline processes multiple speech tasks per subject to generate a single MMSE score prediction.

\subsection{Model Selection Rationale}
A number of important factors led to the choice of Whisper and RoBERTa as our foundation models. OpenAI's Whisper exhibits strong performance under a range of acoustic circumstances and remarkable resilience to various speech patterns and accents. \cite{radford2022whisper}. Given that cognitive impairment frequently presents as speech abnormalities and variations, this resilience is very important for our application.

Because of its improved training approach and exceptional ability to capture subtle linguistic aspects, we selected RoBERTa for text representation over alternative language models. By using dynamic masking, bigger batch sizes, and longer training sequences, RoBERTa enhances BERT's architecture \cite{liu2019roberta}. These gains are especially pertinent to our goal since cognitive evaluation frequently necessitates an awareness of subtle linguistic patterns and possible speech abnormalities that could point to cognitive deterioration. 

\subsection{Data Preprocessing and Normalization}
Our preprocessing pipeline consists of several crucial steps designed to ensure robust model training and evaluation:

\begin{enumerate}
    \item Audio Preprocessing:
    \begin{itemize}
        \item The initial preprocessing stage involves resampling the raw audio recordings to a sampling frequency of 16 kHz, which is essential for optimal performance of the Whisper ASR model as it aligns with the model's pre-training specifications.
    \end{itemize}

    \item Score Normalization:
    For each MMSE score $y$, we compute the normalized score $\hat{y}$ as:
    \begin{equation}
    \hat{y} = \frac{y - y_{\min}}{y_{\max} - y_{\min}}
    \label{eq:normalization}
    \end{equation}
    where $y_{\min}$ and $y_{\max}$ are determined exclusively from the training set. This normalization serves two purposes:
    \begin{itemize}
        \item Stabilizes the training process by bringing all target values into a consistent range
        \item Facilitates better gradient flow through the neural network
    \end{itemize}
\end{enumerate}

\subsection{Speech Processing Pipeline}
The system processes three distinct speech tasks per subject: Cookie Theft Description (CTD), Phonemic Fluency Test (PFT), and Semantic Fluency Test (SFT). Each audio recording $x_i$ undergoes a two-stage transformation process:

\subsubsection{Speech-to-Text Conversion}
Using the Whisper ASR model, we convert each audio recording into text:
\begin{equation}
T_i = \text{Whisper}(x_i)
\label{eq:whisper}
\end{equation}
where $T_i$ represents the transcribed text for the $i$-th recording. Whisper processes the audio in 30-second segments.

\subsubsection{Contextual Feature Extraction}
The transcribed text is processed through RoBERTa to obtain contextual embeddings:
\begin{equation}
h_i = \text{RoBERTa}(T_i)
\label{eq:roberta}
\end{equation}
where $h_i \in \mathbb{R}^{768}$ represents the [CLS] token embedding. This special token accumulates context from the entire sequence through RoBERTa's self-attention mechanisms, effectively creating a fixed-dimensional representation.

\subsection{Model Architecture}
Our architecture consists of two main components:

\subsubsection{Base Model}
The pre-trained RoBERTa model processes tokenized text input:
\begin{equation}
\text{RoBERTa}: \{w_1, ..., w_n\} \rightarrow \mathbb{R}^{768}
\label{eq:encoder}
\end{equation}
where $\{w_1, ..., w_n\}$ represents the token sequence. The model employs 12 transformer layers with 12 attention heads each, processing sequences up to 512 tokens in length.

\subsubsection{Regression Head}
A carefully designed two-layer neural network maps the contextual embeddings to a normalized MMSE score:
\begin{equation}
f(h) = W_2(\text{ReLU}(W_1h + b_1)) + b_2
\label{eq:regression}
\end{equation}
where $W_1 \in \mathbb{R}^{64\times768}$, $W_2 \in \mathbb{R}^{1\times64}$, and $b_1$, $b_2$ are the learnable parameters. The architecture of the regression head was chosen to:
\begin{itemize}
    \item Reduce the high-dimensional RoBERTa embeddings gradually
    \item Introduce non-linearity through ReLU activation
\end{itemize}

\subsection{Training Process}
The model is trained end-to-end using a carefully designed procedure:

\begin{enumerate}
    \item Multi-Task Integration:
    \begin{itemize}
        \item Process all three task recordings (CTD, PFT, SFT) per subject
        \item Generate individual predictions for each task
    \end{itemize}

    \item Loss Computation:
    The model is optimized using Mean Squared Error (MSE) loss:
    \begin{equation}
    \mathcal{L} = \frac{1}{N}\sum_{i=1}^N(\hat{y}_i - y_i)^2
    \label{eq:loss}
    \end{equation}
    where $\hat{y}_i$ is the predicted normalized score and $y_i$ is the ground truth normalized score.

\end{enumerate}

\subsection{Inference and Score Generation}
During inference, the model generates predictions through a two-step process:

\begin{enumerate}
    \item Generate normalized prediction $\hat{p}$
    \item Denormalize to obtain final MMSE score:
    \begin{equation}
    p = \hat{p} \cdot (y_{\max} - y_{\min}) + y_{\min}
    \label{eq:denormalization}
    \end{equation}
\end{enumerate}

Model performance is evaluated using Root Mean Square Error (RMSE) on denormalized predictions:
\begin{equation}
\text{RMSE} = \sqrt{\frac{1}{N}\sum_{i=1}^N(p_i - y_i)^2}
\label{eq:rmse}
\end{equation}
where $p_i$ and $y_i$ are the predicted and actual MMSE scores in the original scale.

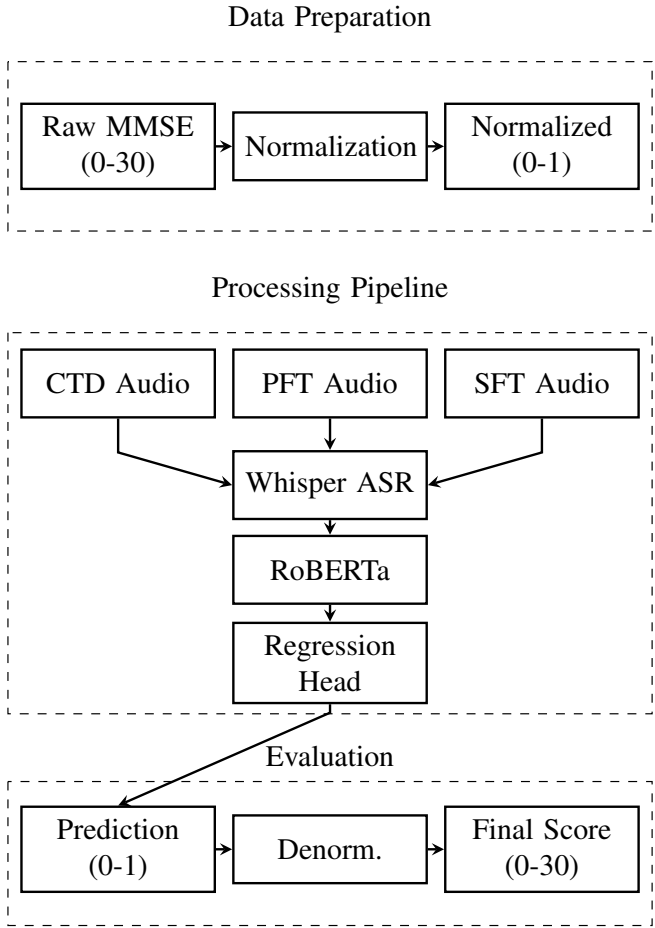
\begin{figure}[t]
\centering
\resizebox{\columnwidth}{!}{
\begin{tikzpicture}[
    auto,
    block/.style={
        rectangle,
        draw=black,
        thick,
        fill=white,
        text width=2cm,
        align=center,
        minimum height=0.8cm,
        inner sep=4pt
    },
    arrow/.style={
        thick,
        ->,
        >=stealth
    }
]

\node[text width=7cm, align=center] at (0,1) {Data Preparation};
\draw[dashed] (-3.8,0.5) rectangle (3.8,-1.5);

\node[block] (raw) at (-2.5,-0.5) {Raw MMSE\\(0-30)};
\node[block] (norm) at (0,-0.5) {Normalization};
\node[block] (scaled) at (2.5,-0.5) {Normalized\\(0-1)};

\node[text width=7cm, align=center] at (0,-2.2) {Processing Pipeline};
\draw[dashed] (-3.8,-2.7) rectangle (3.8,-7.2);

\node[block] (audio1) at (-2.5,-3.3) {CTD Audio};
\node[block] (audio2) at (0,-3.3) {PFT Audio};
\node[block] (audio3) at (2.5,-3.3) {SFT Audio};

\node[block] (whisper) at (0,-4.5) {Whisper ASR};
\node[block] (roberta) at (0,-5.5) {RoBERTa};
\node[block] (regress) at (0,-6.6) {Regression\\Head};

\node[text width=7cm, align=center] at (0,-7.7) {Evaluation};
\draw[dashed] (-3.8,-8.0) rectangle (3.8,-9.7);

\node[block] (pred) at (-2.5,-8.8) {Prediction\\(0-1)};
\node[block] (denorm) at (0,-8.8) {Denorm.};
\node[block] (final) at (2.5,-8.8) {Final Score\\(0-30)};

\draw[arrow] (raw.east) -- (norm.west);
\draw[arrow] (norm.east) -- (scaled.west);

\draw[arrow] (audio1.south) -- +(0,-0.4) -- (whisper.west);
\draw[arrow] (audio2.south) -- (whisper.north);
\draw[arrow] (audio3.south) -- +(0,-0.4) -- (whisper.east);

\draw[arrow] (whisper.south) -- (roberta.north);
\draw[arrow] (roberta.south) -- (regress.north);

\draw[arrow] (regress.south) -- ($(regress.south)+(0,-0.1)$) -- (pred.north);

\draw[arrow] (pred.east) -- (denorm.west);
\draw[arrow] (denorm.east) -- (final.west);

\end{tikzpicture}
}
\caption{System architecture for MMSE score prediction. The pipeline consists of three phases: data preparation (normalization), processing (speech-to-text conversion and feature extraction), and evaluation (prediction and denormalization).}
\label{fig:system_architecture}
\end{figure}
\FloatBarrier

\section{Results}
In this section, we assess three separate models designed to predict Mini-Mental State Examination (MMSE) scores from the speech recordings of patients. These models were trained on a dataset from DementiaBank and the PROCESS challenge, and they were evaluated using different development and test sets. The effectiveness of each model was quantified using RMSE, and the findings are summarized in Table \ref{tab:results}.

\begin{table}[htbp]
\caption{Comparative RMSE Scores for Evaluated Models}
\centering
\resizebox{\columnwidth}{!}{%
\begin{tabular}{|c|c|c|c|}
\hline
\textbf{S No.} & \textbf{Model} & \textbf{Dev RMSE} & \textbf{Test RMSE} \\
\hline
1 & roBERTa (egemaps) & 2.4869 & 2.6911 \\
\hline
2 & SVM (egemaps + Whisper text) & 2.366 & 3.1424 \\
\hline
3 & SVM (egemaps + Whisper text + HP tuning) & 2.411 & 4.1793 \\
\hline
4 & Baselines & - & 2.9850 \\
\hline
\end{tabular}%
}
\label{tab:results}
\end{table}

The regression Model 1 ranked 5th in the regression task in the PROCESS challenge, with its performance falling within a very narrow margin from the other top models.

\section{Experimentation}

The PROCESS Grand Challenge presented us with two tasks:
\begin{enumerate}
    \item \textbf{Classification}: Determining the class of speech samples as dementia or non-dementia.
    \item \textbf{Regression}: Predicting the Mini-Mental State Examination (MMSE) scores, which can subsequently indicate the dementia class the patient belongs to.
\end{enumerate}

The test set results for this challenge were provided by the PROCESS organizers and remained unknown to the participants, ensuring an unbiased evaluation of model performance. In the classification task, multiple models were trained on the DementiaBank and PROCESS Challenge datasets. The F1 scores for the classification models on the development (Dev) and test datasets are summarized in Table \ref{tab:F1_score}

\begin{table}[htbp]
\caption{Comparative F1 Scores for Evaluated Models}
\centering
\resizebox{\columnwidth}{!}{%
\begin{tabular}{|c|c|c|c|}
\hline
\textbf{S No.} & \textbf{Model} & \textbf{Dev F1 score} & \textbf{Test F1 score} \\
\hline
1 & SVM(eGeMAPs + readability) & 0.397 & 0.5193 \\
\hline
2 & Bert(DementiaBank+Process + eGeMAPs) & 0.51 & 0.4876 \\
\hline
3 & (SVM 60\% + BERT 40\%) (DementiaBank+ Process + eGeMAPs) & 0.61 & 0.3998 \\
\hline
4 & Baselines & - & 0.5500 \\
\hline
\end{tabular}%
}
\label{tab:F1_score}
\end{table}
\subsection{Classification Model Methodology}
In the classification task, we sought to improve accuracy by extracting acoustic and linguistic features from the DementiaBank and PROCESS datasets. Model 1 employed a Support Vector Machine (SVM) utilizing eGeMAPS and readability features. Model 2 leveraged BERT, a transformer-based language model, integrated with features from DementiaBank, PROCESS datasets, and eGeMAPS, creating a more complex model. Finally, Model 3 utilized a weighted voting ensemble, combining predictions from SVM and BERT with respective weights of 60

\subsection{Comparison of Model Performance}

Model 1, utilizing SVM with eGeMAPS and readability features, demonstrated better generalization, achieving a higher F1 score on the test dataset (0.5193) compared to its development score. This suggests that the model's simplicity allowed it to adapt well to unseen data. In contrast, Model 2, based on BERT with a broader feature set, performed better on the development dataset (0.510) but showed a drop in test performance (0.4876), likely due to overfitting. Model 3, an ensemble approach using weighted voting, achieved the best development score (0.610) but struggled to generalize, with its test score dropping significantly to 0.3998.

The disparity between development and test results highlights challenges for complex models like BERT and ensembles, they can achieve high performance on training and development data but risk overfitting without sufficient regularization or data diversity. Simpler models, like the SVM in Model 1, can generalize better to unseen data due to their focused feature set. These findings emphasize the need to carefully balance feature richness and model complexity to ensure robust performance in tasks with limited or imbalanced datasets. As the results for classification were close to the baseline F1 score of 0.5500, we decided to proceed with regression for further analysis.

\section{Discussion}
The PROCESS challenge dataset included three types of data: SFT, PFT, and CTD. Here, we provide definitions for these data types:

\subsection{Data Types Description}
\begin{itemize}
    \item \textbf{SFT (Simple Formant Tracking):} This type of data involves basic formant tracking, which is crucial for analyzing vowel sounds in speech. Formants are concentrated bands of acoustic energy and are essential for understanding speech clarity and quality.
    \item \textbf{PFT (Pitch Frequency Tracking):} PFT data focuses on the frequency tracking of pitch in speech. This is particularly important in tone languages, where pitch variations can change meanings, and is also used in detecting emotions through speech tones.
    \item \textbf{CTD (Cookie Theft Detection):} Cookie theft detection in speech processing involves analyzing a person's ability to describe a specific image to assess cognitive functions. CTD was the primary data type used in our models due to its clarity and accuracy, which are imperative for effective training of speech recognition systems.
\end{itemize}

Only the CTD among the three kinds of provided data proved to be useful while we were performing the classification task. The Automatic Speech Recognition (ASR) models trained on this data encountered difficulties due to occasional noises and unclear audio inputs leading to mispronunciations and inaccuracies. Moreover, the use of filler words by speakers, which were often mispronounced by the ASR models, contributed to further inaccuracies and misclassifications.

Also, the models occasionally misclassified healthy control (HC) patients as having MCI or dementia due to their similar performances on the PFT and SFT, which suggests the need for further refinement of feature extraction and classification algorithms to better distinguish between these groups.

\subsection{Impact of Audio Quality on ASR Performance}

These findings show us the importance of audio quality in speech recognition. Mispronunciations and the presence of filler words can significantly affect the performance of ASR systems, leading to errors in data classification. Future work should focus on enhancing the robustness of ASR models to better handle these imperfections in speech inputs.

\subsection{Analysis of Results}
The findings reveal a considerable variation in model performance across both the development and test sets. Model 1, which employs roBERTa with egemaps features, has exhibited a greater capacity to generalize from the development set to the test set, achieving the lowest RMSE of 2.26911 on the test set. This model's impressive results highlight the importance of contextual embeddings in understanding cognitive health through speech signals. 

On the other hand, Models 2 and 3, which incorporated extra textual features from Whisper and underwent further fine-tuning for enhanced performance (Model 3), struggled to retain their effectiveness on the test dataset. Interestingly, Model 3 saw a significant rise in RMSE, indicating a possible overfitting to the development data as a result of extensive parameter tuning.

\subsection{Implications of Results}



The varying performance among the models gives us important insights into the trade-offs between model complexity and generalization ability. The strong performance of Model 1 highlights the importance of contextual embeddings in capturing the linguistic features related to cognitive assessments. This finding indicates that simpler models with effective linguistic processing capabilities can be more efficient and dependable for these types of applications. 

On the other hand, the decline in performance seen in Models 2 and 3 on the test dataset highlights the risks of overfitting in such problem statements. These findings remind us to carefully evaluate the model's complexity and the advantages of simplifying models to improve their generalizability across diverse datasets. 

They also provide a clear direction for future research in this area, suggesting a focus on developing robust models that strike a balance between complexity and consistent performance across different data distributions.

\section{Conclusion}
The findings from this study provide valuable insights into the effectiveness of different computational models in predicting cognitive decline from speech. Future work will focus on refining these models to enhance their predictive accuracy and generalizability across diverse patient populations.

\end{document}